# Unification-Based Glossing*

**Vasileios Hatzivassiloglou**
Department of Computer Science
Columbia University
New York, NY 10027
vh@cs.columbia.edu

**Kevin Knight**
USC/Information Sciences Institute
4676 Admiralty Way
Marina del Rey, CA 90292
knight@isi.edu

## Abstract

We present an approach to syntax-based machine translation that combines unification-style interpretation with statistical processing. This approach enables us to translate any Japanese newspaper article into English, with quality far better than a word-for-word translation. Novel ideas include the use of feature structures to encode word lattices and the use of unification to compose and manipulate lattices. Unification also allows us to specify abstract features that delay target-language synthesis until enough source-language information is assembled. Our statistical component enables us to search efficiently among competing translations and locate those with high English fluency.

## 1 Background

JAPANGLOSS [Knight et al., 1994; 1995] is a project whose goals are to scale up knowledge-based machine translation (KBMT) techniques to handle Japanese-English newspaper MT, to achieve higher quality output than is currently available, and to develop techniques for rapidly constructing MT systems. We built the first version of JAPANGLOSS in nine months and recently participated in an ARPA evaluation of MT quality [White and O'Connell, 1994]. JAPANGLOSS is an effort within the larger PANGLOSS [NMSU/CRL et al., 1995] MT project.

Our approach is to use a KBMT framework, but to fall back on statistical methods when knowledge gaps arise (as they inevitably will). We syntactically analyze Japanese text, map it to a semantic representation, then generate English. Figure 1 shows a sample translation.

Parsing is bottom-up, driven by an augmented context-free grammar whose format is roughly like that of [Shieber, 1986]. Our grammar rules look like this:

```
((NP -> S NP)
  ((X1 syn infl) = (*OR* kihon ta-form rentai))
  ((X0 syn) = (X2 syn))
  ((X0 syn comp) = plus)
  ((X0 syn s-mod) = (X1 syn)))
```

*This work was supported in part by the Advanced Research Projects Agency (Order 8073, Contract MDA904-91-C-5224) and by the Department of Defense.

```
INPUT:

新会社は、2月にも設立される見通し。

INTERLINGUA:

((sem
   ((instance HAVE-AS-A-GOAL)
    (senser <1> ((instance COMPANY-BUSINESS)
                 (q-mod ((instance NEW-VIRGIN)))))
    (phenomenon ((instance FOUND-LAUNCH)
                 (agent <1>)
                 (temporal-locating
                    ((instance MONTH)
                     (index 2))))))))

OUTPUT:

The new company plans to establish in February.
```

Figure 1: Sample JAPANGLOSS Translation.

The semantic representation contains conceptual tokens drawn from the 70,000-term SENSUS ontology [Knight and Luk, 1994]. Semantic analysis proceeds as a bottom-up walk of the parse tree, in the style of Montague and Moore [Dowty et al., 1981; Moore, 1989]. Semantics is compositional, with each parse tree node assigned a meaning based on the meanings of its children. Leaf node meanings are retrieved from a semantic lexicon, while meaning composition rules handle internal nodes. Semantic rules and lexical entries are sensitive to syntactic structure, e.g.:

```
((N -> "kaisha")
  ((x0 sem instance) = COMPANY-BUSINESS))

((NP -> S NP)
  ((X2 syn form) = (*NOT* rentaidome))
  ((X0 sem instance) = rc-modified-object)
  ((X0 sem head) = (X2 sem))
  ((X0 sem rel-mod) = (X1 sem))
  (*OR* (((X1 map subject-role) =c X2))
        (((X1 map object-role)  =c X2))
        (((X1 map object2-role) =c X2))))
```

Generation is performed by PENMAN [Penman, 1989], which includes a large systemic grammar of English. Gaps in the generator's knowledge are filled with statistical techniques [Knight and Hatzivassiloglou, 1995; Knight and Chander, 1994], including a model that can rank potential generator outputs. The English lexicon includes 91,000 roots, comparable in size to the 130,000 roots used in Japanese syntactic analysis.

All of these KBs, however, are still not enough to drive full semantic throughput. Major missing pieces include a large Japanese semantic lexicon and a set of ontological constraints. We are attacking these problems with a combination of manual and automatic techniques [Okumura and Hovy, 1994; Knight and Luk, 1994]. Meanwhile, we want to test our current lexicons, rules, and analyzers in an end-to-end MT system.

We have therefore modified our KBMT system to include a short-cut path from Japanese to English, which we describe in this paper. This path skips semantic analysis and knowledge-based generation, but it uses the same syntactic analyses, lexicons, etc., as the full system. We call this short-cut *glossing*, and it features a new component called the *glosser*, whose job is to transform a Japanese parse tree into English, using easily obtainable resources. Our glosser achieves 100% throughput, even when the parser fails to fully analyze the input sentence and only produces a fragmentary parse tree.

## 2 Bottom-Up Glossing

In thinking about the glossing problem—turning Japanese parse trees into English—we had the following goals and insights:

- Quality. Glossing necessarily involves guessing, as is most obvious from an ambiguous word like *bei*, which may be glossed as either *rice* or *American*. Without a semantic analysis, improved guessing is the road to improved quality.
  1. All potential translation guesses can be packed into an English word lattice, of the sort used in speech recognition systems.
  2. Guesses can be ranked with a statistical language model and the most promising ones can be identified with a search procedure.
- Component re-use. It is possible to build a glosser very quickly if we re-use representations and modules from a full KBMT system.
  1. Word lattices can be stored and manipulated as feature structures.
  2. The compositional semantic interpreter can serve as a glosser, if we provide new knowledge bases.
  3. The statistical model we built for ranking generator outputs [Knight and Hatzivassiloglou, 1995] can also be used for glossing.

This section describes how we put together an MT system based on these ideas. We concentrate here on the components and knowledge bases, deferring linguistic and statistical aspects to following sections.

Word lattices model ambiguities from three sources—Japanese syntactic analysis, lexical glossing, and English synthesis. Here is a small sample lattice:

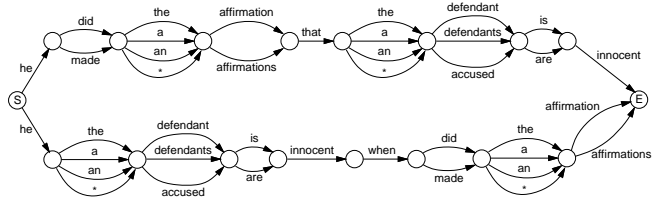

This lattice encodes 768 possible translations; the two main pathways correspond to two different parses. The star symbol (*) stands for an empty transition.

Our original bottom-up semantic analyzer transforms parse trees into semantic feature structures. However, we can make it produce word lattices if we encode lattices with disjunctive feature structures, e.g.:

```
((gloss
  (*OR*
   ((op1 "he")
    (op2 (*OR* "did" "made"))
    (op3 ((op1 (*OR* "the" "a" "an" "*empty*"))
          (op2 (*OR* "affirmation" "affirmations"))
          (op3 "that")
          (op4 ((op1 ((op1 (*OR* "the" "a"
                                 "an" "*empty*"))
                      (op2 (*OR* "defendant"
                                 "defendants"
                                 "accused"))))
                (op2 ((op1 (*OR* "is" "are"))
                      (op2 "innocent")))))))
   ((op1 "he")
    (op2 (*OR* "defendant" "defendants" "accused"))
    (op3 ((op1 (*OR* "is" "are"))
          (op2 "innocent")))
    (op4 "when")
    (op5 ((op1 (*OR* "did" "made"))
          (op2 ((op1 (*OR* "the" "a" "an"
                           "*empty*"))
                (op2 (*OR* "affirmation"
                           "affirmations"))))))))))
```

In the above representation, `*OR*` marks mutually disjoint components of the gloss, while the features `op1`, `op2`, etc. represent sequentially ordered portions of the gloss.

This structure can be transformed automatically into a format suitable for statistical processing. As part of that transformation, we also do a bit of English morphology, to simplify the analyzer's work. The analyzer still runs as a bottom-up walk of the parse tree, using unification to implement Montague-style composition. However, we replace the conventional semantic lexicon with a gloss lexicon, easily obtainable from an online Japanese-English dictionary:

```
((N -> "kaisha")
 ((x0 gloss) = (*OR* "company" "firm")))
```

We also replace semantic rules with glossing rules, e.g.:

```
((NP -> S NP)
  ((X0 gloss op1) = (X2 gloss))
  ((X0 gloss op2) = (*OR* "which" "that"))
  ((X0 gloss op3) = (X1 gloss))
  ((X0 tmp) = (X2 tmp)))
```

This rule says: to gloss a Japanese noun phrase (NP) created from a relative clause (S) combining with another noun phrase (NP), glue together the following—an English gloss of the child NP, a relative pronoun (either *which* or *that*), and an English gloss of the S. The rule also propagates abstract features (tmp) from the child NP to the parent. We return to these features in the next section.

We built a set of 171 complex rules to match the structures in our syntactic grammar. Our new semantic analyzer composes glosses (word lattices) rather than meanings, so we call it the *glosser*.

Figure 2 compares glossing and semantic interpretation. Each parse tree node is annotated with its analysis. Sentence-level analyses appear at the top. These analyses are then fed to subsequent JAPANGLOSS modules—to the generator (in the case of semantic analysis) or directly to the statistical model (in the case of glossing).

## 3 Linguistic Aspects

As Figure 2 shows, semantic interpretation makes much more flexible use of unification as a combinator than glossing does. In fact, most of our glossing rules simply concatenate word lattices and insert function words. Concatenation lets us put a direct object after a verb in English, for example, even though it comes before the verb in Japanese. However, many Japanese structures are different enough from English that this strategy breaks down. Consider the sentence *John ga Bill ni tabesaseta*, parsed as:

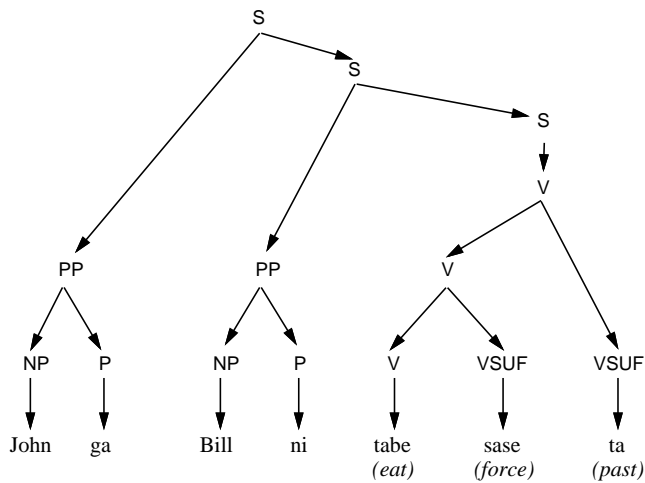

One translation of this sentence into English is *John forced Bill to eat*. The difficulty is how to assign word lattices to intermediate nodes in the parse tree. If we assign *forced to eat* to *tabesaseta*, there will be no way for us to squeeze in the word *Bill* at the next level. Our solution is to use unification to pass abstract features up the parse tree. We store abstract information under a top-level feature called tmp, parallel to the gloss and syn (for syntactic) feature structures. So the feature structure at the lowest S node looks like:

```
((gloss (*OR* "eat" "ingest"))
 (tmp ((force +) (past +))))
```

The complex (S -> PP S) rule then successfully unpacks the abstract features into words at the next level:

```
((gloss ((op1 "forced")
         (op2 "Bill")
         (op3 "to")
         (op4 (*OR* "eat" "ingest")))))
```

How to efficiently turn bundles of abstract features into English is a difficult general problem lying at the heart of natural language generation. Our glosser tackles only simple instances of this problem, involving at most three of four features. Three binary features can require up to eight rules to "spell out," and we sometimes must specify all cases. Often, however, we see a decomposition in which one abstract feature spells itself out independently of the others. In these cases, there is no exponential blowup in the required number of glosser rules.

Here is a fragment of rules dealing with the above example:

```
((V -> V VSUF)
  ((x0 tmp) = (x1 tmp))
  (*XOR*
    (((x2 syn entry-form) = "sase")
     ((x0 gloss) = (x1 gloss))
     ((x0 tmp force) = +))
    (((x2 syn entry-form) = "ta")
     ((x1 tmp force) =c +)
     ((x0 gloss) = (x1 gloss))
     ((x0 tmp past) = +))
    (((x2 syn entry-form) = "ta")
     ((x0 gloss op1) = (x1 gloss))
     ((x0 gloss op2) = "+past"))))

((S -> V)
  ((x0 gloss) = (x1 gloss))
  ((x0 tmp) =  (x1 tmp)))

((PP -> NP P)
  ((x0 syn entry-form) = (x2 syn entry-form))
  ((x0  gloss)  = (x1 gloss)))

((S -> PP S)
  (*XOR*
    (((x1 syn entry-form) = "ga")
     ((x0 gloss op1) = (x1 gloss))
     ((x0 gloss op2) = (x2 gloss)))
    (((x1 syn entry-form) = "ni")
     ((x2 tmp force) =c +)
     (*XOR* (((x2 tmp past) =c +)
             ((x0 gloss op1) = "forced"))
            (((x0 gloss op1) =
                (*OR* "force" "forces"))))
     ((x0 gloss op2) = (x1 gloss))
     ((x0 gloss op3) = "to")
     ((x0 gloss op4) = (x2 gloss)))))
```

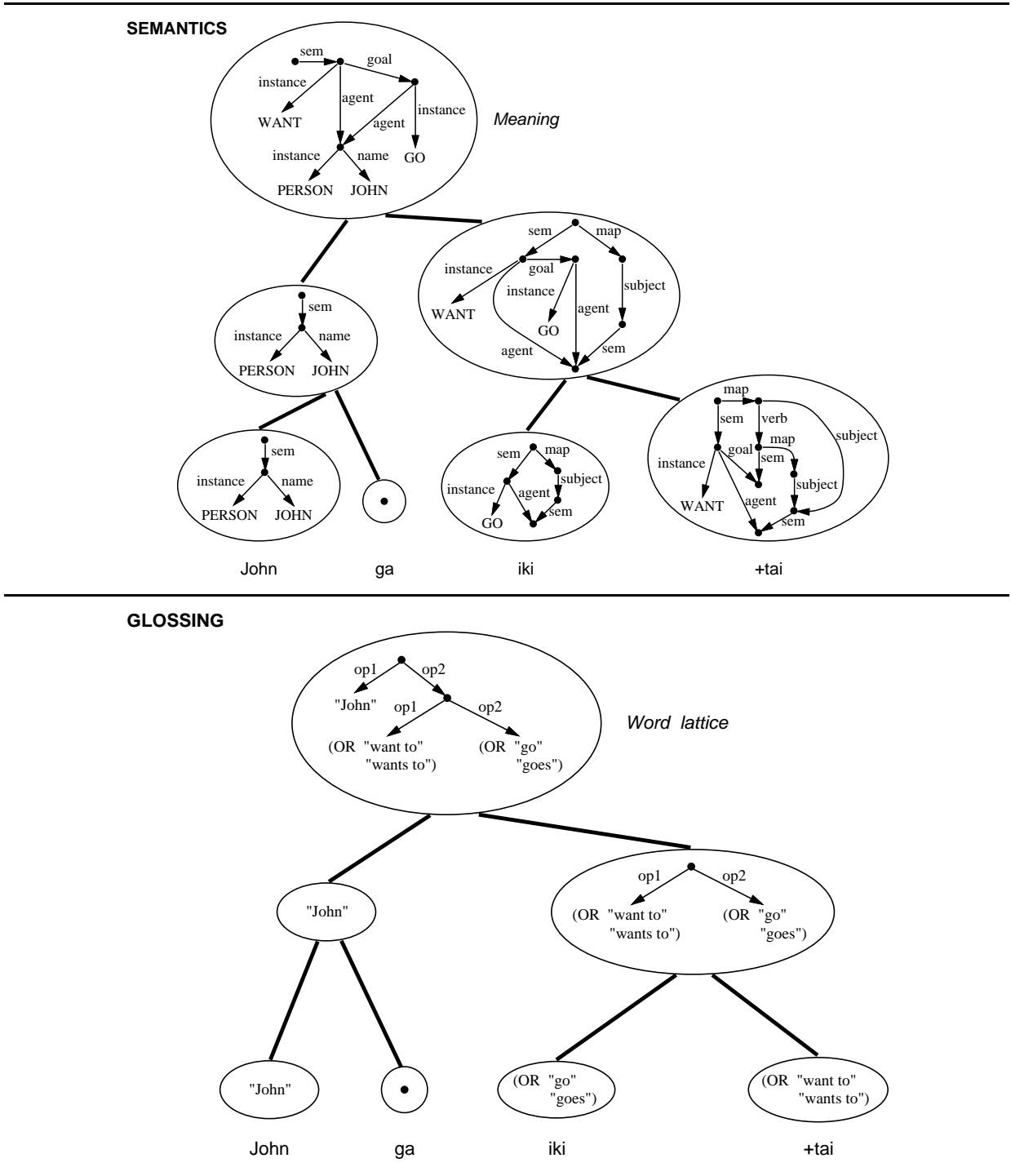

Figure 2: Semantic analysis versus glossing. Both convert parse trees into feature structures using unification-based compositional techniques. Semantics computes a conceptual representation while glossing computes a target language word lattice.

In this notation, =c (a symbol borrowed from Lexical-Functional Grammar [Kaplan and Bresnan, 1982]) means the feature sequence must already exist in the incoming child constituent, and *XOR* sets up a disjunction of feature constraints, only one of which is allowed to be satisfied.

## 4 Statistical Language Modeling

Our glosser module proposes a number of possible translations for each Japanese word or sequence of words that have been matched to a syntactic constituent by the bottom-up parser. Each such translation unit represents a *lexical island* according to the knowledge the glosser has, i.e., a piece of text for which no other constraints are available. At the same time, the various renditions of each translation unit can combine together, leading to many possible translations at the sentence level. In order to select among the many combinations of these possibilities, we need an objective function that will score them, hopefully ranking the correct translation near the top. To accomplish this task, we approximate correctness by fluency, and further approximate fluency by likelihood, selecting the combination of words and phrases that seems most likely to occur in the target language. This approach offers two advantages:

- Because we measure likelihood at the sentence level, we take into account interactions between words and phrases that are produced from different parts of the Japanese input. For example, *bei* in Japanese may mean either *American* or *rice*, and *sha* may mean either *shrine* or *company*. If both possibilities survive for both words after the glosser processes *beisha*, the likelihood criterion will select *American company*, which is almost always the correct translation. In addition, ranking potential translations by their probability in the target language indirectly handles collocational constraints and allows the correct choice of function words which may not appear in the source text at all (e.g., articles in Japanese) or are subject to non-compositional lexical constraints (e.g., prepositions in English, as in *afraid of*, or *on Monday* versus *in February*).

- In the absence of additional lexical constraints originating from neighboring target language words and phrases, individual translations containing more common and widely used words are preferred over translations that contain more rare and obscure words. In this way, the Japanese word *kuruma* will be translated as *car* and not as *motorcar*. This tactic is optimal when no disambiguating information is available, since it selects the most likely translation, avoiding rare and very specialized alternatives.

In the remainder of this section we discuss how we measure the probability of an English sentence from the probabilities of short sequences of words (n-grams [Bahl et al., 1983]), how we estimate these basic probabilities of n-grams, how we handle problems of sparse data by smoothing our estimates, and how we search the space of translation possibilities efficiently during translation so as to select the best scoring translations.

### 4.1 The sentence likelihood model

As we discussed in the previous paragraph, we want to associate with each English sentence $A$ a likelihood measure $\Pr(A)$. Since the number of such sequences is very large and our training text is not unlimited, we cannot expect to count the occurrences of $A$ in a corpus and then use a classic estimation technique such as maximum likelihood estimation.[1] Instead, we adopt a Markov assumption, according to which the probability of seeing a given word depends only on the short history of words appearing just before it in the sentence. Using a history of one or two previous words, the stochastic process that generates sequences of English words is approximated by a first or second order Markov chain (bigram or trigram model) respectively. For reasons of numerical accuracy with finite precision computations, we convert probabilities to log-likelihoods. Then, the log-likelihood of a sequence of words $S = w_1 w_2 \ldots w_n$ is

$$\mathrm{LL}(S) = \sum_i \log \Pr(w_i | w_{i-1}) \qquad \text{for bigrams}$$

$$\mathrm{LL}(S) = \sum_i \log \Pr(w_i | w_{i-1}, w_{i-2}) \qquad \text{for trigrams}$$

Unfortunately, this likelihood model will assign smaller and smaller probabilities as the sequence becomes longer. Since we need to compare alternative translations of different lengths, we alleviate this problem by adding a heuristic corrective bonus which is an increasing function of sentence length. After experimenting with several such functions, we have found that the function $f(n) = 0.5n$, where $n$ is the length of the word sequence, gives satisfactory results when added to the log-likelihood measure. This is equivalent to adding an exponential function of length to the original probabilities.

### 4.2 Estimating n-gram probabilities

To estimate the conditional bigram and trigram probabilities used in our model, we processed a large corpus of carefully written English texts and we measured the frequencies of one-, two-, and three-word sequences in it. Since we aim at translation of unrestricted Japanese newspaper articles, we selected the Wall Street Journal (WSJ) corpus[2] as the most representative available collection of English texts that our output should imitate. We processed the 1987 and 1988 years from the WSJ corpus, giving us 46 million words of training text, containing approximately 300,000 different word types.

The large number of different word types makes our modeling task significantly more complicated than previous similar language models. These models were usually designed for speech recognition tasks, where the vocabulary was limited to at most a few thousand frequent English words. With our vocabulary of 300,000 words, we have $9 \cdot 10^{10}$ different bigrams, and $2.7 \cdot 10^{16}$ different trigrams. Handling such large numbers of n-grams is problematic in terms of storage space and retrieval speed. Furthermore, estimating probabilities for these is difficult, since most of them do not occur in our training text.[3]

---

[1] Even with unlimited text, such an approach is not feasible because of practical limitations in terms of memory and hardware speed.

[2] Available from the ACL Data Collection Initiative as CD Rom 1.

In order to reduce the number of n-grams for which we need to estimate probabilities, we first implemented a simple schema of class-based smoothing. We developed finite-state automata that use features like word position, capitalization, and types of characters in the word to separate the words into one of four classes: numbers, monetary amounts, proper names, and regular words. We then treat all words in each of the first three classes as the same word, pooling their frequencies together and using uniform maximum likelihood estimates for all words in each class, irrespective of whether the particular word has been seen in the training corpus or not.

The class-based smoothing reduces the number of words for which we need to estimate individual probabilities to 120,000; more importantly, it reduces the number of bigrams by a factor of 7.5 and the number of trigrams by a factor of 22.5. Still, many of the surviving n-grams have not been observed in the training corpus, and to estimate their probability as zero would be clearly incorrect, given the compositionality of the English language. Good [1953] has proposed a method that addresses this problem and is theoretically optimal under rather general distributional assumptions (namely, that each n-gram follows a marginal binomial distribution). The resulting Good-Turing estimator replaces the observed frequency $r$ with the corrected frequency

$$r^* = (r+1)\frac{N_{r+1}}{N_r}$$

where $N_r$ is the number of n-grams that occur $r$ times. The corrected frequencies $r^*$ are subsequently used to provide estimates of probabilities through the maximum likelihood formula. In general, probability mass is "stolen" from the observed n-grams (proportionally more from those that have been observed a few times than from the more frequent ones), and redistributed to the unseen n-grams.

The Good-Turing estimator still suffers from one disadvantage; it assigns the same probability to all n-grams that have not been seen in the corpus (and for that matter, to all n-grams that have been seen the same number of times in the corpus). Church and Gale [1991] have proposed an *enhanced* version of the estimator for bigrams, in which a secondary predictor based on unigram (word) probabilities is used to separate the bigrams into *bins*, and the Good-Turing formula is applied separately within each bin. The rationale of this approach is that pairs of frequent (or infrequent) words are expected to be frequent (or infrequent) themselves; departures from this norm become notable when the bigrams are separated into bins according to the likelihood of their component words, and these differences lead to different estimates for the bigram probabilities even if the bigrams have the same frequency in the corpus. Church and Gale [1991] provide empirical evidence that indicates that the enhanced Good-Turing estimator outperforms both simple estimators such as the MLE and (to a lesser extent) other complex estimators such as an enhanced version of the deleted estimation method [Jelinek and Mercer, 1980].

We have implemented the basic Good-Turing method for single words, allowing for 130,000 unseen words. We then use the Good-Turing estimates of probabilities of words to compute the secondary predictor $\log(\Pr(A)\Pr(B))$, in the enhanced Good Turing estimator for the bigram $AB$. We have also extended the enhanced method to trigrams $ABC$, using the secondary predictor $\log(\Pr(AB)\Pr(C))$ that combines the estimated probabilities of the initial bigram and the final word.[4] We smooth the secondary predictor by using 2 or 3 bins per order of magnitude (for trigrams and bigrams, respectively), and we smooth the counts of n-grams in each bin ($N_r$) with a dynamically self-adjusting local smoother.

### 4.3 Searching the word lattice space

In the previous two subsections we discussed how any sequence of words can be assigned a likelihood estimate, appropriately modified for length. This in principle would allow ranking the various translation alternatives by simply computing the likelihood for each of them. However, the word lattices produced by the glosser compactly encode billions of possible translations: a simple linear chain of states with 30 states and two arcs leaving each state represents $2^{30}$, or about one billion, paths, each corresponding to a potential translation.

Consequently, a method is needed to efficiently search the word lattice and select a small set of highly likely translations. We adopted the *N-best* algorithm for this purpose [Chow and Schwartz, 1989]. Unlike the widely used Viterbi algorithm [Viterbi, 1967], which only produces a single best scoring path, this algorithm offers the advantage of producing any number of the highest scoring paths in the lattice; these paths can then be rescored with a more extensive (and expensive) method. It also offers controlled accuracy (i.e., the extent of suboptimality can be arbitrarily decreased by the amount of memory made available to the search), and empirical studies [Nguyen *et al.*, 1994] have shown that it performs equally well with other, more complicated methods. Finally, no forward estimates of the viability of a partial path are required (as is, for example, the case in the A* or stack decoder [Jelinek *et al.*, 1975; Bahl *et al.*, 1983] algorithm).

We first perform a topological sort of the states in the word lattice, so that we can visit each state after all its predecessors have been processed. As we process a given state, we keep a list of the best scoring sequences of words reaching that state from the start state of the lattice. At each state, we extend all word sequences ending at the predecessors the the current state, recompute their scores, and prune the search space by keeping only a prespecified number of sequences, specified by the width of the global search *beam*. In practice, we have found that a beam of 1,000 hypotheses per node gives accurate results at reasonable search speed. The sequences are stored compactly via pointers to the preceding states (along with information about the specific arc taken at each step), and maintained in a fast priority queue to avoid sorting. This allows us to simulate an HMM of any order, as well as trace any number of final sentences (up to the beam width) when the final state of the lattice is reached. The complexity of the search algorithm is slightly superlinear in terms of the beam width, the number of states, the n-gram length used in the model, and the average fan-out in the lattice (number of arcs leaving each state).

---

[3]Recall that the latter contains $4.6 \cdot 10^7$ words, and consequently only a slightly higher number of bigrams and trigrams, when the special *end-of-sentence* token is taken into account.

[4]$\log(\Pr(A)\Pr(B)\Pr(C))$ is another predictor which we believe may outperform $\log(\Pr(AB)\Pr(C))$ because it is less correlated with $\log(\Pr(ABC))$. But computing this alternative predictor, even off-line, is impractical because of the very large number of possible trigrams.

## 5 Results

The glosser is currently being used in our machine translation system as a fall-back component in cases of parsing or semantic transfer failures. We participated in the most recent (September 1994) ARPA evaluation of machine translation systems (see [White and O'Connell, 1994] for a discussion of the evaluation methodology employed) with promising results. A sample of translations produced by the glossing module is given below, in the form of Japanese input followed by the correct translation and the translation given by the glosser. Due to space limitations, we are showing output on small example sentences, although JAPANGLOSS typically operates on much longer sentences characteristic of newspaper text.

彼は英語にずばぬけた才能を持っている。
*He has unusual ability in English.*
```
He holds a talent that exceeded the English
language.
```

その情報は広まった。
*The news got abroad.*
```
The information spread.
```

生き物は環境の変化に適応できなければならない。
*Living creatures must be adaptable to environmental change.*
```
Animal circumstances accommodation variation
does not enable.
```

日本に来る観光客はきまって富士山がすばらしいと言う。
*Visitors to Japan always admire Mt. Fuji.*
```
Tourists that coming in Japan be decided, and
say that Mt. Fuji is splendid.
```

彼は暴力に反対である。
*He is adverse to violence.*
```
He has the contrary to violence.
```

All the above translations have been obtained with the bigram language model, as heavy computational and storage demands have delayed the deployment of the more precise trigram model. We expect higher quality output when the trigram model becomes fully operational.

## 6 Related Work and Discussion

The glosser described in this paper is a type of transfer MT, and it follows in the tradition of syntax-based MT systems like SYSTRAN. However, our use of statistics allowed us to avoid much of the traditional hand-coding, and to produce a competitive MT system in nine months. Other statistical approaches to MT include CANDIDE [Brown *et al.*, 1993], which does not do a syntactic analysis of the source text, and LINGSTAT [Yamron *et al.*, 1994], which does probabilistic parsing. Both LINGSTAT and JAPANGLOSS require syntax because they translate between languages with radically different word orders.

Our use of features in syntax, glossing, and semantics gives us the flexibility to correct translation errors, capture generalizations, and rapidly build up a complete MT system. As the features become more abstract, the analysis deepens, and our translations improve—this is knowledge-based work. Improvements will also come from better statistical modeling. Our future work will be directed at finding these improvements, and at studying the interaction between knowledge bases and statistics.

## Acknowledgments

We would like to thank Yolanda Gil, Kenji Yamada, and the IJCAI reviewers for helpful comments on a draft of this paper.

## References

[Bahl *et al.*, 1983] L. R. Bahl, F. Jelinek, and R. L. Mercer. A maximum likelihood approach to continuous speech recognition. *IEEE Transactions on Pattern Analysis and Machine Intelligence PAMI*, 5(2), 1983.

[Brown *et al.*, 1993] P. F. Brown, S. A. Della-Pietra, V. J. Della-Pietra, and R. L. Mercer. The mathematics of statistical machine translation: Parameter estimation. *Computational Linguistics*, 19(2), June 1993.

[Chow and Schwartz, 1989] Y. Chow and R. Schwartz. The N-Best algorithm: An efficient search procedure for finding top N sentence hypotheses. In *Proc. DARPA Speech and Natural Language Workshop*, pages 199–202, 1989.

[Church and Gale, 1991] K. W. Church and W. A. Gale. A comparison of the enhanced Good-Turing and deleted estimation methods for estimating probabilities of English bigrams. *Computer Speech and Language*, 5, 1991.

[Dowty *et al.*, 1981] D. R. Dowty, R. Wall, and S. Peters. *Introduction to Montague Semantics*. Reidel, Dordrecht, 1981.

[Good, 1953] I. J. Good. The population frequencies of species and the estimation of population parameters. *Biometrika*, 40, 1953.

[Jelinek and Mercer, 1980] F. Jelinek and R. L. Mercer. Interpolated estimation of Markov source parameters from sparse data. In *Pattern Recognition in Practice*. North-Holland, Amsterdam, 1980.

[Jelinek et al., 1975] F. Jelinek, L. R. Bahl, and R. L. Mercer. Design of a linguistic statistical decoder for the recognition of continuous speech. *IEEE Transactions on Information Theory*, 21(3), 1975.

[Kaplan and Bresnan, 1982] R.M. Kaplan and J. Bresnan. Lexical-functional grammar: A formal system for grammatical representation. In *The Mental Representation of Grammatical Relations*. MIT Press, Cambridge, MA, 1982.

[Knight and Chander, 1994] K. Knight and I. Chander. Automated postediting of documents. In *Proc. AAAI*, 1994.

[Knight and Hatzivassiloglou, 1995] K. Knight and V. Hatzivassiloglou. Two-level, many-paths generation. In *Proc. ACL*, 1995.

[Knight and Luk, 1994] K. Knight and S. K. Luk. Building a large-scale knowledge base for machine translation. In *Proc. AAAI*, 1994.

[Knight et al., 1994] K. Knight, I. Chander, M. Haines, V. Hatzivassiloglou, E. Hovy, M. Iida, S. K. Luk, A. Okumura, R. Whitney, and K. Yamada. Integrating knowledge bases and statistics in MT. In *Proc. Conference of the Association for Machine Translation in the Americas (AMTA)*, 1994.

[Knight et al., 1995] K. Knight, I. Chander, M. Haines, V. Hatzivassiloglou, E. Hovy, M. Iida, S. K. Luk, R. Whitney, and K. Yamada. Filling knowledge gaps in a broad-coverage MT system. In *Proc. IJCAI*, 1995.

[Moore, 1989] R. Moore. Unification-based semantic interpretation. In *Proc. ACL*, 1989.

[Nguyen et al., 1994] L. Nguyen, R. Schwartz, Y. Zhao, and G. Zavaliagkos. Is N-Best dead? In *Proc. ARPA Human Language Technology Workshop*. Advanced Research Projects Agency, 1994.

[NMSU/CRL et al., 1995] NMSU/CRL, CMU/CMT, and USC/ISI. The Pangloss Mark III machine translation system. Technical Report CMU-CMT-95-145, Carnegie Mellon University, 1995. Jointly issued by Computing Research Laboratory (New Mexico State University), Center for Machine Translation (Carnegie Mellon University), Information Sciences Institute (University of Southern California). Edited by S. Nirenburg.

[Okumura and Hovy, 1994] A. Okumura and E. Hovy. Ontology concept association using a bilingual dictionary. In *Proc. of the ARPA Human Language Technology Workshop*, 1994.

[Penman, 1989] Penman. The Penman documentation. Technical report, USC/Information Sciences Institute, 1989.

[Shieber, 1986] S. Shieber. *An Introduction to Unification-Based Approaches to Grammar*. University of Chicago, 1986. Also, CSLI Lecture Notes Series.

[Viterbi, 1967] A. J. Viterbi. Error bounds for convolution codes and an asymptotically optimal decoding algorithm. *IEEE Transactions on Information Theory*, 13:260–269, 1967.

[White and O'Connell, 1994] J. White and T. O'Connell. Evaluation in the ARPA machine translation program: 1993 methodology. In *Proc. ARPA Human Language Technology Workshop*, 1994.

[Yamron et al., 1994] J. Yamron, J. Cant, A. Demedts, T. Dietzel, and Y. Ito. The automatic component of the LINGSTAT machine-aided translation system. In *Proc. ARPA Workshop on Human Language Technology*, 1994.